\documentclass[final,3p,margin=1in]{elsarticle}

\usepackage{amssymb}
\usepackage{amsmath}
\usepackage{amsthm}
\usepackage{booktabs}
\usepackage{multicol}
\usepackage{multirow}
\usepackage{makecell}
\usepackage{algorithm, algorithmic}
\usepackage[utf8]{inputenc}
\usepackage{hyperref}

\usepackage{subcaption}
\usepackage{booktabs}
\usepackage{multirow}
\usepackage{caption}
\usepackage{rotating} 

\makeatletter
\def\bstctlcite#1{\@bsphack
  \@for\@citeb:=#1\do{%
    \edef\@citeb{\expandafter\@firstofone\@citeb}%
    \if@filesw\immediate\write\@auxout{\string\citation{\@citeb}}\fi}%
  \@esphack}
\makeatother

\DeclareUnicodeCharacter{0229}{\c{e}}

\theoremstyle{plain}

\newtheorem*{lemma*}{Lemma}

\theoremstyle{definition}

\numberwithin{theorem}{section}
\numberwithin{definition}{section}
\numberwithin{lemma}{section}
\numberwithin{proposition}{section}
\numberwithin{corollary}{section}
\numberwithin{notation}{section}
\numberwithin{remark}{section}
\numberwithin{example}{section}

\begin{document}
\bstctlcite{IEEEexample:BSTcontrol}
\begin{frontmatter}

\title{PCA-Enhanced Adaptive NVAR Framework for High-Resolution Sea Surface Temperature Forecasting}

\author[label1]{Sherkhon Azimov\corref{cor1}}
\ead{sherxonazimov94@pusan.ac.kr}
\author[label1,label2,label3]{Susana López-Moreno}
\author[label3,label4]{Eric Dolores-Cuenca}
\author[label5]{JinYong Choi}
\author[label1,label6]{Sangil Kim\corref{cor1}}
\ead{sangil.kim@pusan.ac.kr}

\affiliation[label1]{organization={Department of Mathematics, Pusan National University},
            country={Republic of Korea}}

\affiliation[label2]{organization={Humanoid Olfactory Display Center, Pusan National University},
            country={Republic of Korea}}
            
\affiliation[label3]{organization={Industrial Mathematics Center, Pusan National University},
            country={Republic of Korea}}
            
\affiliation[label4]{organization={Department of Mathematics, Yonsei University},
            country={Republic of Korea}}

\affiliation[label5]{organization={Marine Natural Disaster Research Department, Korea Institute of Ocean Science and Technology}, 
country={Republic of Korea}}

\affiliation[label6]{organization={Institute for Future Earth, Pusan National University}, country={Republic of Korea}}
            
\cortext[cor1]{Corresponding authors}

\begin{abstract}
\small

Accurate forecasting of sea surface temperature (SST) is essential for marine ecosystem monitoring, climate assessment, fisheries management, and operational ocean forecasting. While numerical ocean models provide reliable predictions, they are computationally expensive, and conventional machine learning methods often suffer from high-dimensional inputs and error accumulation during long-term autonomous forecasting. This study extends our previously proposed Adaptive Nonlinear Vector Autoregression (Adaptive NVAR) framework to high-resolution real-world SST prediction by integrating Principal Component Analysis (PCA) through Singular Value Decomposition (SVD). Daily SST fields from the GLORYS12V1 reanalysis dataset covering the East Sea, Yellow Sea, and East China Sea are compressed into a lower-dimensional latent representation that preserves the dominant spatial variability. The proposed reduced-order framework is evaluated using autonomous rolling forecasts up to a 90-day horizon and compared with Standard NVAR (Next Generation Reservoir Computing) and a Persistence baseline. Across all three regions, Adaptive NVAR consistently suppresses long-term error accumulation, achieving up to 96.52 \% improvement in mean squared error relative to Persistence while maintaining stable predictive performance throughout extended forecasting. Although the adaptive architecture incurs a higher one-time offline optimization cost than Standard NVAR, inference is completed in milliseconds, making the proposed framework an efficient and scalable approach for long-term, high-resolution ocean state forecasting.
\end{abstract}


\end{frontmatter}

\section{Introduction}

Accurate forecasting of high-resolution Sea Surface Temperature (SST) is essential for monitoring marine ecosystems, assessing climate variability, managing fisheries, supporting maritime transportation, and enabling operational ocean forecasting \cite{ju2022impacts, zhang2021sea}. Regional seas surrounding the Korean Peninsula, including the East Sea, Yellow Sea, and East China Sea, exhibit diverse oceanographic characteristics ranging from energetic mesoscale circulation to shallow shelf dynamics, making them an ideal testbed for evaluating data-driven forecasting methods.

Traditional numerical ocean circulation models provide physically consistent and reliable forecasts by explicitly solving the governing equations of ocean dynamics. However, their high computational cost limits their practicality for real-time forecasting, large ensemble simulations, and applications requiring rapid prediction updates.

To address these limitations, data-driven approaches and machine learning techniques have emerged as attractive alternatives. Nevertheless, forecasting high-resolution SST remains challenging because oceanographic observations are inherently high-dimensional, strongly nonlinear, and exhibit complex spatio-temporal dependencies. Deep learning models have demonstrated promising predictive capabilities, but they often require large computational resources, offer limited physical interpretability, and are susceptible to autoregressive error accumulation during long-term forecasting.

These challenges become particularly evident in the regional seas surrounding the Korean Peninsula, each of which exhibits distinct oceanographic characteristics. The East Sea is dominated by strong mesoscale circulation, where the East Korea Warm Current interacts with colder northern waters to form the highly variable Subpolar Front, generating abundant eddies and pronounced spatial variability \cite{kim2010seasonal, lee2006intermediate, trusenkova2009dynamically}. In contrast, the Yellow Sea is a shallow continental shelf strongly influenced by tidal mixing and seasonal atmospheric forcing, while the East China Sea is shaped by the interaction between the Kuroshio Current and continental shelf circulation. These contrasting dynamical environments give rise to diverse SST evolution patterns, providing a challenging testbed for evaluating the robustness and generalizability of data-driven forecasting methods.

Recently, Next-Generation Reservoir Computing (NG-RC), also widely referred to as the standard Nonlinear Vector Autoregression (standard NVAR) framework, has gained considerable attention as a stable and efficient paradigm for modeling complex dynamical systems  \cite{gauthier2021next}. By leveraging time-delay embeddings and fixed polynomial feature mappings, standard NVAR eliminates the need for expensive recurrent network training while preserving the structural advantages of traditional Reservoir Computing (RC). Comparative studies on chaotic time series have demonstrated that RC techniques can match or even exceed the trajectory-tracking accuracy of Recurrent Neural Networks (RNNs) and Long Short-Term Memory (LSTM) while offering dramatically lower computational overhead and avoiding gradient-based training anomalies \cite{shahi2022prediction}. Despite these benefits, standard NVAR can struggle when deployed on highly nonstationary geophysical processes, where fixed, predetermined feature mappings fail to capture evolving nonlinear interactions over extended periods.

To address these limitations, the Adaptive NVAR framework was recently introduced \cite{sherkhon2025adaptive}. As a data-adaptive nonlinear vector autoregressive model, it utilizes a trainable Multi-Layer Perceptron (MLP) to learn optimal nonlinear feature representations directly from data. This modification drastically improves scalability and performance under high-dimensional, noisy, and complex conditions compared to standard NVAR, Echo State Networks (ESNs), and hybrid ESN architectures, all while preserving a lightweight and elegant structure. The competitive edge of this approach was recently highlighted by Ren et al. \cite{ren2026dctfm} in their development of the Dynamic Causal-Temporal Forecasting Model (DCTFM); when benchmarked against an Adaptive NVAR configuration, the two models yielded highly competitive performance results, with Adaptive NVAR achieving remarkably close accuracy despite its significantly simpler architecture and lower number of parameters. 

This study builds directly upon these advancements, extending the application of the Adaptive NVAR framework from low-dimensional synthetic chaotic benchmarks to real-world geophysical forecasting challenges. To handle the high spatial dimensionality inherent to oceanographic fields, we present a unified spatiotemporal forecasting framework that merges Singular Value Decomposition (SVD)—analogous to spatial Principal Component Analysis (PCA)—with the Adaptive NVAR model to predict high-resolution SST dynamics in the East Sea.

The structure of our proposed approach is as follows:
\begin{itemize}
    \item \textbf{Dimensionality Reduction:} High-resolution spatial SST fields are compressed into a lower dimensional space using SVD to identify the main empirical orthogonal functions that influence regional ocean dynamics.
    \item \textbf{Latent-Space Dynamics Modeling:} We model the temporal changes of these latent states using the Adaptive NVAR framework. This framework uses a multi-layer perceptron (MLP) to learn optimal nonlinear feature representations dynamically instead of depending on static polynomial combinations.
    \item \textbf{Spatiotemporal Reconstruction:} We map the predicted latent trajectories back to the original spatial domain using an inverse PCA transformation, resulting in full-resolution, multi-step-ahead spatial SST forecasts.
\end{itemize}

We assess the performance of our framework by using the high-resolution Copernicus Marine Service Global Ocean Physics Reanalysis (GLORYS12V1) dataset  \cite{jean2021copernicus}. Our results show that the Adaptive NVAR model consistently outperforms standard NVAR/NG-RC setups. It achieves significantly lower error metrics across various forecasting periods of up to 90 days. Additionally, incorporating SVD-based dimensionality reduction significantly reduces the computational requirements. This creates a scalable and fast framework ideal for real-time oceanographic applications.

The rest of this paper is organized as follows: Section \ref{Sec:Methodology} explains the theoretical basis of the standard and adaptive NVAR frameworks along with the dimensionality reduction process. Section \ref{Sec:Numerical experiments} describes the numerical experiments, dataset features, hyperparameter tuning, and forecasting results. Finally, Section \ref{Sec:Conclusion} presents the conclusions of the paper, outlines its limitations, and explores directions for future research.

\section{Methodology}
\label{Sec:Methodology}

\subsection{Spatiotemporal Dimensionality Reduction via Singular Value Decomposition}
\label{subsec:svd_reduction}

To efficiently model high-dimensional geophysical fields, we use a principal component analysis (PCA) based on the SVD to map the spatial data into a low-dimensional, orthogonal latent space \cite{hannachi2007empirical}. Edward Lorenz was one of the first to introduce this mathematical tool to atmospheric and climate science \cite{lorenz1956empirical}. Let $\mathbf{U} \in \mathbb{R}^{M \times T}$ represent the spatiotemporal data matrix. Here, $M$ is the total number of spatial grid points on a surface layer, and $T$ is the number of temporal snapshots. The full SVD of the data matrix is expressed as:
\begin{equation*}
    \mathbf{U} = \mathbf{\Phi} \mathbf{\Sigma} \mathbf{V}^\top,
\end{equation*}
where $\mathbf{\Phi} \in \mathbb{R}^{M \times M}$ contains the orthogonal spatial eigenvectors (empirical orthogonal functions, or EOFs), $\mathbf{\Sigma} \in \mathbb{R}^{M \times T}$ is a rectangular diagonal matrix containing the singular values in descending order, and $\mathbf{V} \in \mathbb{R}^{T \times T}$ represents the temporal coefficients.

While large climate datasets with significant dimensions often need probabilistic, randomized SVD methods \cite{halko2011finding} to prevent memory issues, the smaller spatial scale of the East Sea area allowed us to perform a complete, exact deterministic decomposition.
To simplify the structure and remove high-frequency spatial noise, we create a low-dimensional approximation by truncating the expansion to the first $r$ dominant modes ($r \ll M$):
\begin{equation*}
    \mathbf{U} \approx \mathbf{\Phi}_r \mathbf{\Sigma}_r \mathbf{V}_r^\top = \mathbf{\Phi}_r \mathbf{Z},
\end{equation*}
where $\mathbf{\Phi}_r \in \mathbb{R}^{M \times r}$ denotes the truncated spatial basis matrix, and $\mathbf{Z} = \mathbf{\Sigma}_r \mathbf{V}_r^\top \in \mathbb{R}^{r \times T}$ represents the reduced-order temporal state trajectories. The mathematical optimality of this low-rank approximation is guaranteed by the Eckart-Young-Mirsky theorem \cite{eckart1936approximation}, which proves that the truncated SVD minimizes the reconstruction error in terms of both the Frobenius and spectral matrix norms.
The matrix $\mathbf{Z}$ serves as the direct low-dimensional input time series $\mathbf{u}_t$ for both the standard NVAR baseline (Section \ref{subsec:standard_nvar}) and the proposed Adaptive NVAR framework (Section \ref{subsec:adaptive_nvar}). Following autoregressive state prediction in the latent space, the physical fields are reconstructed via linear combination with the preserved spatial modes:
\begin{equation*}
    \hat{\mathbf{u}}_{t+1}^{\mathrm{physical}} = \mathbf{\Phi}_r \hat{\mathbf{u}}_{t+1}.
\end{equation*}

\subsection{Next-Generation Reservoir Computing}
\label{subsec:standard_nvar}

The standard nonlinear vector autoregression (NVAR) framework, serving as the core mathematical formulation for Next-Generation Reservoir Computing (NG-RC) \cite{gauthier2021next}, completely replaces the random, recurrent internal spaces of traditional reservoir networks with an explicit, deterministic spatiotemporal feature mapping. Let $\mathbf{u}_t \in \mathbb{R}^{d}$ represent the $d$-dimensional input vector at time step $t$. The total NVAR feature vector $\mathbf{O}_t$ is constructed via concatenation:
\begin{equation*}
    \mathbf{O}_t = c \oplus \mathbf{O}_t^{\mathrm{lin}} \oplus \mathbf{O}_t^{\mathrm{nonlin}},
\end{equation*}
where $c$ is a constant scalar bias term and $\oplus$ denotes vector concatenation. The delay-embedded linear feature vector $\mathbf{O}_t^{\mathrm{lin}}$ at time step $t$ is explicitly structured as \cite{kantz2003nonlinear}:
\begin{equation*}
    \mathbf{O}_t^{\mathrm{lin}} = \mathbf{u}_t \oplus \mathbf{u}_{t-\tau} \oplus \mathbf{u}_{t-2\tau} \oplus \dots \oplus \mathbf{u}_{t-(k-1)\tau},
\end{equation*}
where $k$ is the delay embedding length parameter and $\tau$ is the delay stride spacing. Consequently, the linear feature vector $\mathbf{O}_t^{\mathrm{lin}}$ spans a total dimension of $k \cdot d$. 

A core structural advantage of this formulation over traditional recurrent architectures is that NVAR does not require an extended warm-up (or washout) period to clear transient initial states; it requires exactly $k_s = (k-1)\tau + 1$ historical steps to establish its operational feature space \cite{gauthier2021next}. For simplicity and structural consistency across models, we set $\tau=1$ for all NVAR variations evaluated in this work.

The nonlinear feature component $\mathbf{O}_t^{\mathrm{nonlin}}$ is built directly from the unique elements of the linear delay vector. Following the standard NG-RC convention \cite{gauthier2021next}, a polynomial quadratic expansion is applied, meaning $\mathbf{O}_t^{\mathrm{nonlin}}$ contains all unique second-order monomials formed from the elements of $\mathbf{O}_t^{\mathrm{lin}}$:
\begin{equation*}
    \mathbf{O}_t^{\mathrm{nonlin}} = \mathbb{V}\left( \mathbf{O}_t^{\mathrm{lin}} \otimes \mathbf{O}_t^{\mathrm{lin}} \right),
\end{equation*}
where $\otimes$ denotes the vector outer product and $\mathbb{V}(\cdot)$ extracts the $N_{\mathrm{nonlin}} = \frac{1}{2}(k d)(k d + 1)$ unique, non-redundant terms. 

\begin{figure}[ht]
    \centering
    \includegraphics[width=0.7\linewidth]{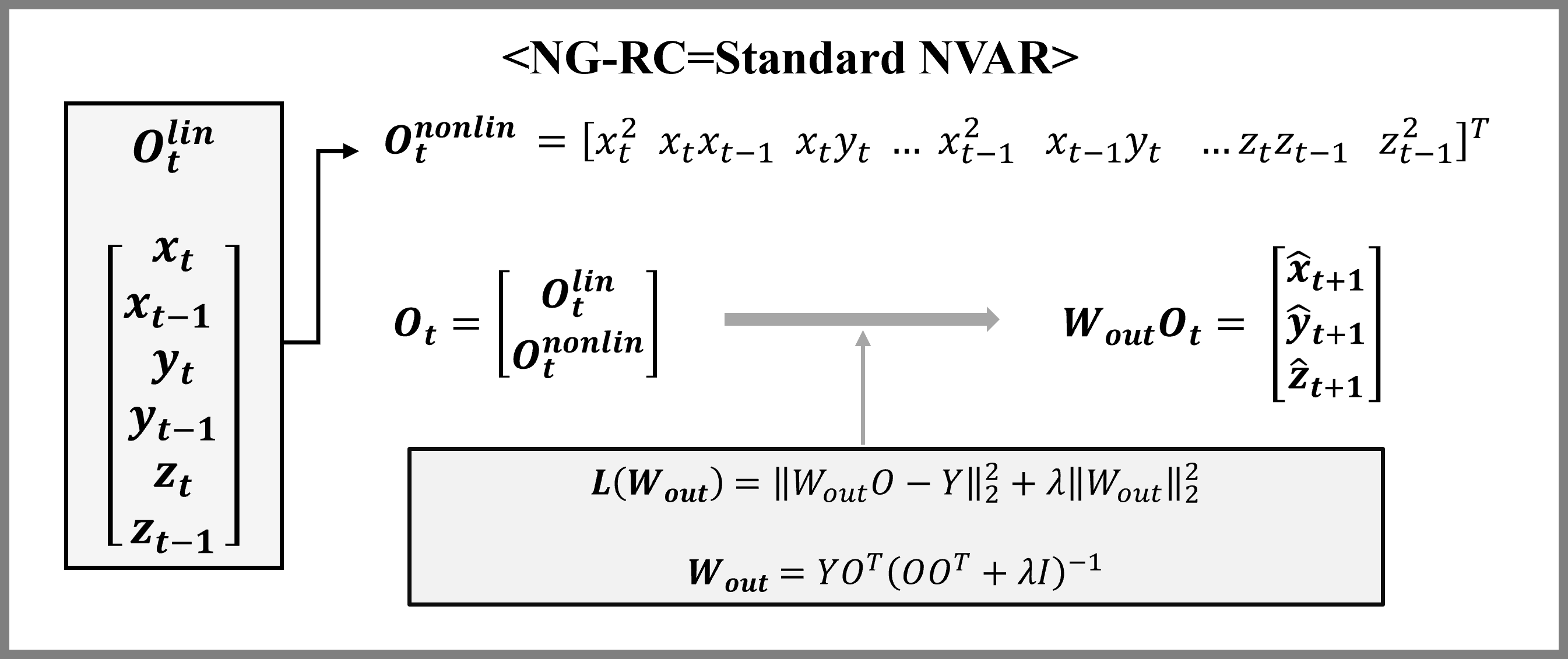}
    \caption{Schematic representation of the Next-Generation Reservoir Computing (or standard NVAR) framework showing deterministic delay embedding and polynomial feature generation.}
    \label{fig:nvar_schematic}
\end{figure}

The optimization of the NVAR framework utilizes a Tikhonov regularized least-squares cost function, resulting in the closed-form formulation \cite{tikhonov1977solutions}:
\begin{equation*}
    \mathbf{W}_{\mathrm{out}} = \mathbf{Y} \mathbf{O}^\top \left( \mathbf{O} \mathbf{O}^\top + \lambda \mathbf{I} \right)^{-1},
\end{equation*}
where $\mathbf{O}$ is the assembled feature matrix across all training points, $\lambda > 0$ is the regularization parameter, and $\mathbf{I}$ is the identity matrix. Following \cite{gauthier2021next}, the target labels are formulated as forward operational differences, $\mathbf{Y}_t = \mathbf{u}_{t+1} - \mathbf{u}_t$, yielding an Euler-like integration rollout rule for the autoregressive forecasting:
\begin{equation}
    \hat{\mathbf{Y}}_t = \mathbf{W}_{\mathrm{out}} \mathbf{O}_t.
    \label{eq:nvar_euler_prediction}
\end{equation}
This difference-based target is preserved across all variations to ensure comparative consistency. A schematic representation of the standard NVAR pipeline is illustrated in Figure~\ref{fig:nvar_schematic}.

\subsection{Adaptive Nonlinear Autoregressive Model}
\label{subsec:adaptive_nvar}

 While standard NVAR and Adaptive NVAR share an identical linear delay-embedding vector $\mathbf{O}_t^{\mathrm{lin}}$ and an incremental Euler prediction rule, they differ fundamentally in how their nonlinear feature spaces are formed and optimized. 

Instead of relying on hardcoded, static polynomial features, Adaptive NVAR handles complex spatiotemporal structures by modeling nonlinear feature spaces using a shallow, trainable multi-layer perceptron (MLP) \cite{sherkhon2025adaptive}. Leveraging the properties of neural networks as universal function approximators \cite{hornik1989multilayer}, we introduce a parameterized mapping $\phi_{\mathbf{\Theta}}: \mathbb{R}^{kd} \rightarrow \mathbb{R}^{m}$ that transforms the delay embedding space dynamically:
\begin{equation*}
    \mathbf{O}_t^{\mathrm{nonlin}} = \phi_{\mathbf{\Theta}}\left( \mathbf{O}_t^{\mathrm{lin}} \right),
\end{equation*}
where $\mathbf{\Theta}$ denotes the total set of learnable network weights and biases, and $m$ specifies the output feature dimensionality produced by the network. In practice, this neural mapping is realized via a shallow MLP architecture formulated as:
\begin{equation}
    \phi_{\mathbf{\Theta}}\left( \mathbf{O}_t^{\mathrm{lin}} \right) = \sigma\left( \mathbf{W}_2 \cdot \sigma\left( \mathbf{W}_1 \mathbf{O}_t^{\mathrm{lin}} + \mathbf{b}_1 \right) + \mathbf{b}_2 \right),
    \label{eq:mlp_structure}
\end{equation}
where $\mathbf{W}_1, \mathbf{W}_2, \mathbf{b}_1, \mathbf{b}_2 \in \mathbf{\Theta}$ represent the trainable structural parameters, and $\sigma(\cdot)$ denotes an element-wise nonlinear activation function. 

The resulting total feature space is formed by concatenating the original linear delay-embedded vector directly with the learned nonlinear representation via an explicit structural skip connection:
\begin{equation}
    \mathbf{O}_t^{\mathrm{total}} = \mathbf{O}_t^{\mathrm{lin}} \oplus \mathbf{O}_t^{\mathrm{nonlin}}.
    \label{eq:adaptive_total_concat}
\end{equation}
Note that an explicit constant bias term is omitted from \eqref{eq:adaptive_total_concat}, since bias vectors are implicitly handled within the neural network parameters $\mathbf{\Theta}$.

The temporal rollout update uses a linear readout layer that operates over the combined feature space to predict the state increment, matching the target structure in \eqref{eq:nvar_euler_prediction}:
\begin{equation*}
    \hat{\mathbf{Y}}_t = \mathbf{W}_{\mathrm{out}} \mathbf{O}_t^{\mathrm{total}} = \mathbf{W}_{\mathrm{out}} \left( \mathbf{O}_t^{\mathrm{lin}} \oplus \phi_{\mathbf{\Theta}}\left(\mathbf{O}_t^{\mathrm{lin}}\right) \right).
\end{equation*}
The operational flow of this skip-connection-based forward pass is detailed systematically in Algorithm~\ref{alg:adaptive_nvar_forward}.
\begin{figure}[h]
    \centering
    \includegraphics[width=0.7\linewidth]{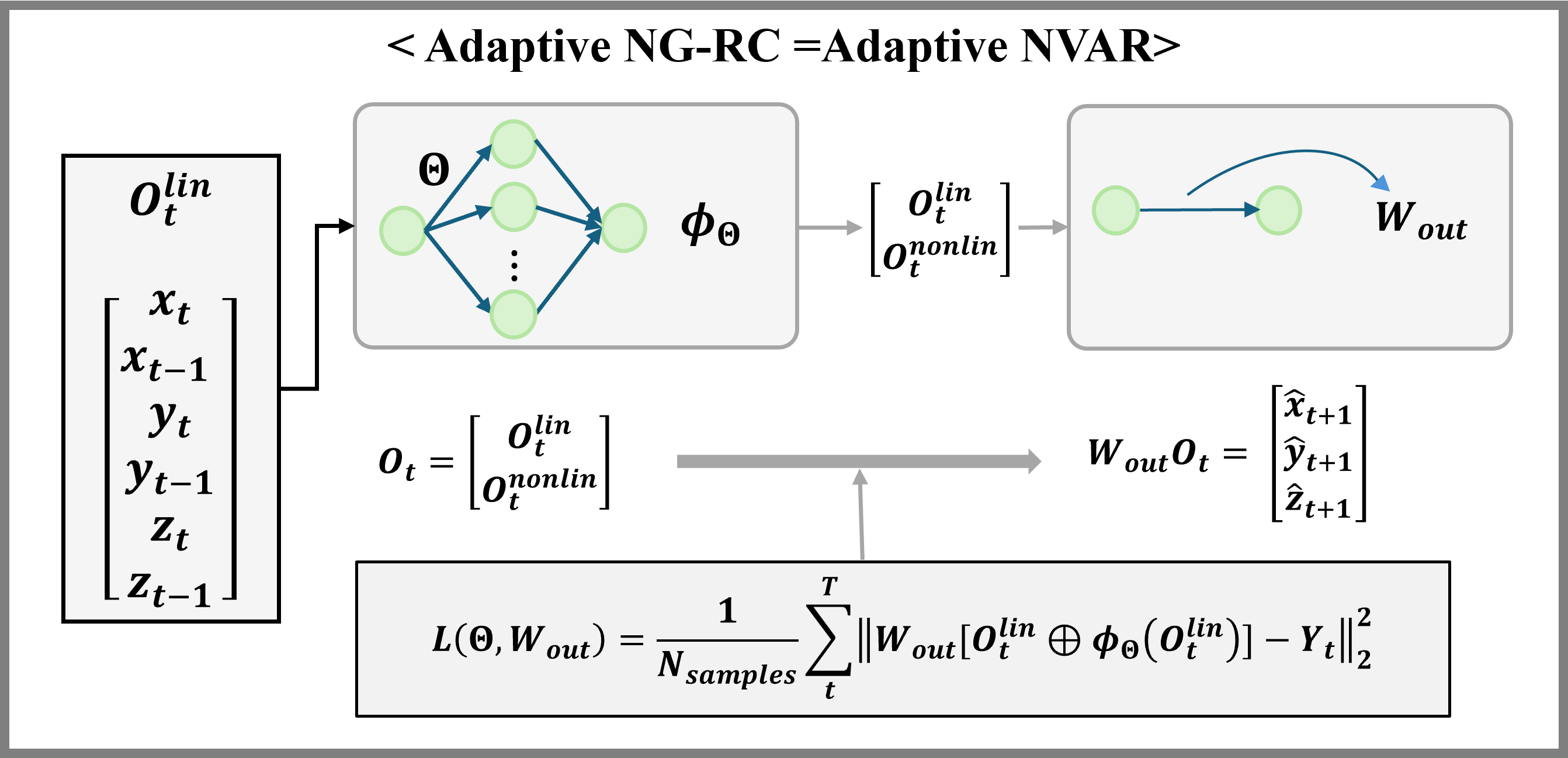}
    \caption{Schematic representation of the Adaptive NVAR framework showcasing the learnable MLP nonlinear mapping and the unified skip-connection optimization architecture.}
    \label{fig:adaptive_nvar_schematic}
\end{figure}

\begin{algorithm}[ht]
\caption{Adaptive NVAR}
\label{alg:adaptive_nvar_forward}
\begin{algorithmic}[1]
    \REQUIRE Linear delay-embedded trajectory vector $\mathbf{O}_t^{\mathrm{lin}}$
    \ENSURE Predicted state increments $\hat{\mathbf{Y}}_t$
    \STATE \textbf{Parameters:} Trainable neural mapping parameters $\mathbf{\Theta}$, readout weights $\mathbf{W}_{\mathrm{out}}$
    \STATE \textbf{Nonlinear Feature Generation:} Compute $\mathbf{O}_t^{\mathrm{nonlin}} \leftarrow \phi_{\mathbf{\Theta}}\left(\mathbf{O}_t^{\mathrm{lin}}\right)$ via \eqref{eq:mlp_structure}
    \STATE \textbf{Skip-Connection:} Form combined feature space $\mathbf{O}_t^{\mathrm{total}} \leftarrow \mathbf{O}_t^{\mathrm{lin}} \oplus \mathbf{O}_t^{\mathrm{nonlin}}$
    \STATE \textbf{Readout Evaluation:} Compute predicted increment $\hat{\mathbf{y}}_t \leftarrow \mathbf{W}_{\mathrm{out}} \mathbf{O}_t^{\mathrm{total}}$
    \RETURN $\hat{\mathbf{y}}_t$
\end{algorithmic}
\end{algorithm}

Unlike standard configurations that rely on static matrix inversions, Adaptive NVAR trains its parameters end-to-end. The linear readout weights $\mathbf{W}_{\mathrm{out}}$ and internal network parameters $\mathbf{\Theta}$ are optimized jointly using a gradient-based optimization procedure that minimizes the mean squared error over the valid training indices:
\begin{equation*}
    \mathcal{L}_{\mathrm{Adaptive}} = \frac{1}{N_{\mathrm{samples}}} \sum_{t=k_s}^{T} \left\|  \mathbf{W}_{\mathrm{out}} \left( \mathbf{O}_t^{\mathrm{lin}} \oplus \phi_{\mathbf{\Theta}}\left(\mathbf{O}_t^{\mathrm{lin}}\right) \right) -\mathbf{Y}_t \right\|_2^2,
\end{equation*}
where $\mathbf{Y}_t = \mathbf{u}_{t+1} - \mathbf{u}_t$ is the true observed state increment matrix and $N_{\mathrm{samples}}$ is the total number of training steps. This joint optimization forces the extracted nonlinear features to adapt dynamically to the true underlying system trajectories, eliminating the need for fixed hand-engineered feature matrices and improving stability over long horizons. An overview of the proposed framework is shown in Figure~\ref{fig:adaptive_nvar_schematic}.

\section{Numerical Experiments}
\label{Sec:Numerical experiments}

\subsection{Dataset Description}

We evaluate the proposed framework using the Copernicus Marine Service Global Ocean Physics Reanalysis (GLORYS12V1) dataset \cite{jean2021copernicus}. Driven by the NEMO ocean circulation model and assimilating satellite and \textit{in situ} observations, GLORYS12V1 provides daily oceanographic state estimates on a $1/12^\circ$ horizontal grid (approximately 8~km) across 50 vertical levels from 1993 onward. For all forecasting experiments, we extract sea surface temperature (SST) at the uppermost layer (0.49~m depth).

\begin{figure}[h!]
\centering
\includegraphics[width=\linewidth]{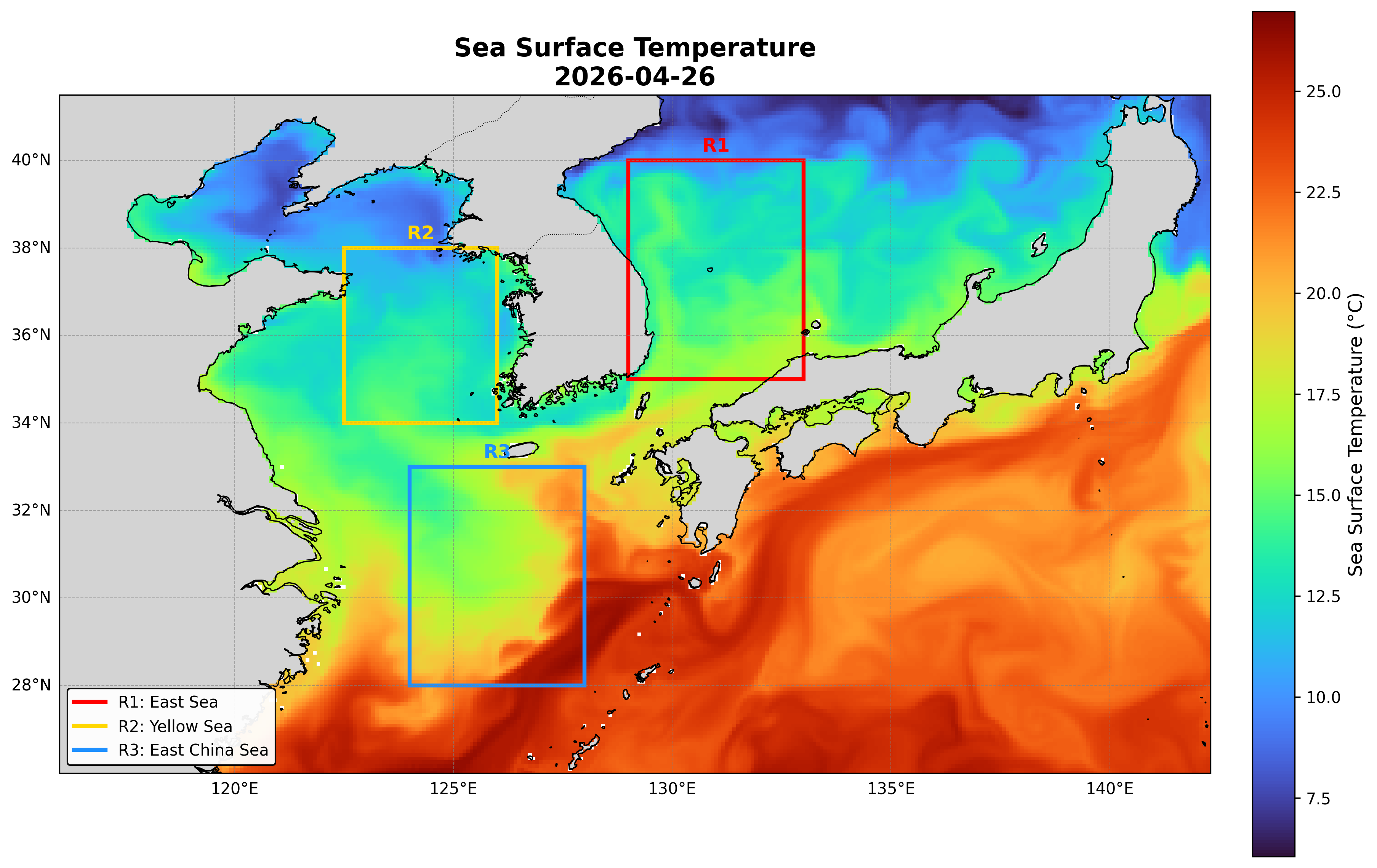}
\caption{Geographic coverage of the three study regions overlaid on the long-term mean SST: R1 (East Sea), R2 (Yellow Sea), and R3 (East China Sea).}
\label{fig:sst_snapshot}
\end{figure}

To test the Adaptive NVAR framework across contrasting oceanographic regimes, we selected three representative domains surrounding the Korean Peninsula: the East Sea (R1), the Yellow Sea (R2), and the East China Sea (R3). Figure~\ref{fig:sst_snapshot} displays the geographic boundaries of these domains, and Table~\ref{tab:study_regions} summarizes their spatial dimensions and grid characteristics.

\begin{table}[h!]
\centering
\caption{Geographic domains and dataset characteristics for the three study regions.}
\label{tab:study_regions}
\begin{tabular}{lcccc}
\hline
\textbf{Region} & \textbf{Latitude} & \textbf{Longitude} & \textbf{Grid Size} & \textbf{Ocean Cells} \\
\hline
R1: East Sea       & $35^\circ$--$40^\circ$N   & $129^\circ$--$133^\circ$E   & $61 \times 49$ & 2,801 \\
R2: Yellow Sea     & $34^\circ$--$38^\circ$N   & $122.5^\circ$--$126^\circ$E & $49 \times 43$ & 2,078 \\
R3: East China Sea & $28^\circ$--$33^\circ$N   & $124^\circ$--$128^\circ$E   & $61 \times 49$ & 2,989 \\
\hline
\end{tabular}
\end{table}

For each domain, we retrieved 5,000 continuous daily SST snapshots spanning August 19, 2012, through April 27, 2026. The raw field tensor for each region has shape $(5000,\,1,\,H,\,W)$, where $H$ and $W$ denote the latitude and longitude grid dimensions listed in Table~\ref{tab:study_regions}. Land grid cells (represented as NaNs) were masked out, and then flattened into a vector of $d$ active ocean cells. Stacking these vectors temporally yields a 2D feature matrix
\[
X \in \mathbb{R}^{T \times d},
\]
where $T = 5000$ time steps and $d$ represents the number of valid ocean spatial locations. These matrices serve as the raw inputs for dimensionality reduction and subsequent time-series modeling.

All source data were obtained in NetCDF-4 format directly from the Copernicus Marine Service platform.

\subsection{Dimensionality Reduction}

Even after masking land points, the flattened feature matrices $X$ remain high-dimensional, presenting a computational bottleneck for dynamic modeling. To extract the primary spatial modes and lower the system's dimensionality, we perform principal component analysis (PCA) via singular value decomposition (SVD) on each region independently.

To prevent data leakage into validation and testing phases, the PCA transformation matrix was constructed using only the initial 4,000 time steps.

\begin{figure}[t]
\centering
\includegraphics[width=\linewidth]{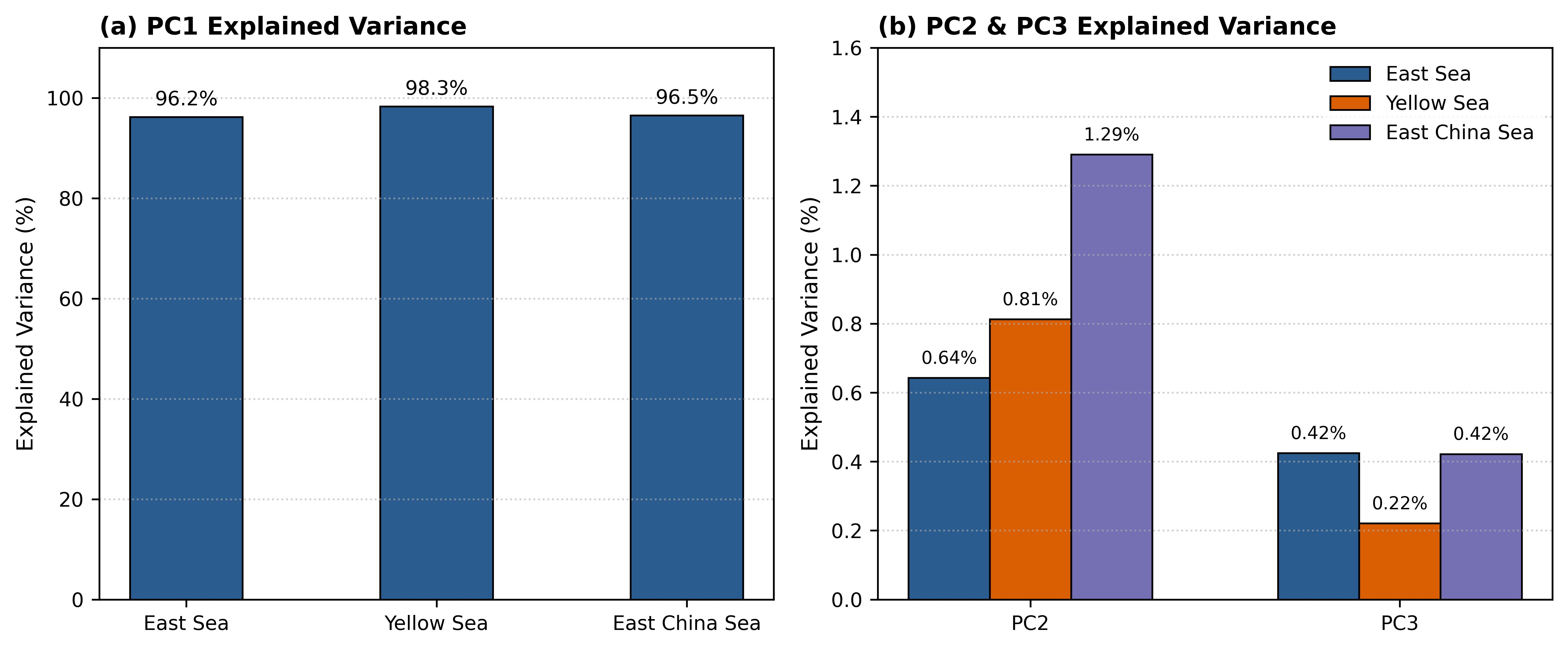}
\caption{Proportion of explained variance captured by the first three principal components in each study region.}
\label{fig:pca_variance}
\end{figure}

As shown in Figure~\ref{fig:pca_variance}, the first principal component (PC1) accounts for almost all variance across all three regions—explaining 96.19\% in the East Sea, 98.29\% in the Yellow Sea, and 96.51\% in the East China Sea. By contrast, PC2 and PC3 collectively contribute less than 2.0\% of the total variance, confirming that the spatial SST variability in these domains is strongly governed by a single dominant mode.

Accordingly, we retained only the leading mode ($n_{\mathrm{comp}} = 1$) across all numerical experiments. This one-dimensional latent trajectory acts as the temporal state variable for both standard NVAR and Adaptive NVAR. Once predictions are generated in the latent space, full 2D spatial SST fields are reconstructed by applying the inverse PCA transformation.

\subsection{Training and Evaluation Procedure}

The 5,000-snapshot time series was split chronologically into warm-up/training, validation, and testing periods following an 80/10/10 ratio. Specifically, the sequence comprises 100 warm-up steps, 3,900 training steps, 500 validation steps, and 500 testing steps. Models were evaluated on lead-time prediction horizons of 10, 30, 60, and 90 days.

To quantify forecasting performance, we compute three standard error metrics across $N$ spatial grid points and a forecast horizon of $T$ time steps. Let $y_{i,t}$ and $\hat{y}_{i,t}$ denote the observed and predicted SST values at spatial location $i$ and lead time $t$, respectively:

\begin{itemize}
    \item \textbf{Mean Squared Error (MSE):}
    \[
    \mathrm{MSE} = \frac{1}{NT}\sum_{i=1}^{N}\sum_{t=1}^{T}\left(y_{i,t}-\hat{y}_{i,t}\right)^2
    \]
    \item \textbf{Mean Absolute Error (MAE):}
    \[
    \mathrm{MAE} = \frac{1}{NT}\sum_{i=1}^{N}\sum_{t=1}^{T}\left|y_{i,t}-\hat{y}_{i,t}\right|
    \]
    \item \textbf{Root Mean Squared Error (RMSE):}
    \[
    \mathrm{RMSE} = \sqrt{\frac{1}{NT}\sum_{i=1}^{N}\sum_{t=1}^{T}\left(y_{i,t}-\hat{y}_{i,t}\right)^2}
    \]
\end{itemize}

To obtain robust performance estimates across the test period, we apply a rolling-window evaluation protocol. The final reported error for each metric represents the mean across all $W$ rolling forecast windows:
\[
\overline{\mathrm{MSE}} = \frac{1}{W} \sum_{w=1}^{W} \mathrm{MSE}^{(w)}, \quad
\overline{\mathrm{MAE}} = \frac{1}{W} \sum_{w=1}^{W} \mathrm{MAE}^{(w)}, \quad
\overline{\mathrm{RMSE}} = \frac{1}{W} \sum_{w=1}^{W} \mathrm{RMSE}^{(w)}.
\]

For parameter optimization, the Adaptive NVAR framework was trained using the Adam optimizer \cite{kingma2014adam}, providing fast and stable convergence before its performance was evaluated using full multi-step rolling forecasts.

\subsection{Hyperparameter Optimization}
\label{Subsec:Hyperparameter Optimization}

To ensure a fair and rigorous comparison between Standard NVAR and the proposed Adaptive NVAR framework, hyperparameter tuning was conducted independently for each study region using the Tree-structured Parzen Estimator (TPE) algorithm in Optuna \cite{akiba2019optuna, bergstra2011algorithms}. Optimization was performed over 200 trials to minimize the mean RMSE of the rolling windows at the maximum forecast horizon ($t = 90$ days). To account for stochastic weight initialization and ensure stable performance metrics, each trial for Adaptive NVAR evaluated the mean performance of 5 independent model runs.

The search space encompassed key structural, architectural, and regularization parameters, as detailed in Table~\ref{tab:hyperparameter_search}.

\begin{table}[h!]
\centering
\caption{Hyperparameter search space evaluated during TPE optimization.}
\label{tab:hyperparameter_search}
\renewcommand{\arraystretch}{1.2}
\begin{tabular}{llc}
\hline
\textbf{Model} & \textbf{Hyperparameter} & \textbf{Search Space} \\
\hline
Standard NVAR 
& Delay embedding length ($k$) & $\{2, 5, 10, 20, 30, 40, 50\}$ \\
& Ridge regularization ($\lambda$) & $[10^{-7}, 10^{6}]$ \\
\hline
Adaptive NVAR 
& Delay embedding length ($k$) & $\{2, 5, 10, 20, 30, 40, 50\}$ \\
& Number of hidden units ($d_h$) & $\{20, 50, 100\}$ \\
& Initial Adam learning rate ($\eta_{\mathrm{Adam}}$) & $[10^{-4}, 10^{-1}]$ \\
\hline
\end{tabular}
\end{table}

Model validation and testing were carried out across overlapping rolling forecast windows with a 90-day horizon and a 1-day stride, generating 411 autonomous forecasting evaluation windows per domain. Table~\ref{tab:optimal_configurations} presents the resulting optimal hyperparameter configurations and total optimization times across all three study regions.

\begin{table}[h!]
\centering
\caption{Optimal hyperparameter configurations and optimization times across study regions.}
\label{tab:optimal_configurations}
\renewcommand{\arraystretch}{1.25}
\setlength{\tabcolsep}{4.5pt}
\small

\begin{tabular}{lccccc}
\hline
\textbf{Model / Region} & \textbf{$k$} & \textbf{\begin{tabular}[c]{@{}c@{}}Hidden\\ Dim. ($d_h$)\end{tabular}} & \textbf{\begin{tabular}[c]{@{}c@{}}Adam LR\\ ($\eta_{\mathrm{Adam}}$)\end{tabular}} & \textbf{\begin{tabular}[c]{@{}c@{}}Ridge\\ ($\lambda$)\end{tabular}} & \textbf{\begin{tabular}[c]{@{}c@{}}Execution\\ Time (h)\end{tabular}} \\
\hline
\multicolumn{6}{l}{\textit{Adaptive NVAR}} \\
R1: East Sea       & 50 & 100 & $5.50 \times 10^{-2}$ & -- & 4.18 \\
R2: Yellow Sea     & 50 & 100 & $4.22 \times 10^{-2}$ & -- & 4.10 \\
R3: East China Sea & 50 & 100 & $4.15 \times 10^{-2}$ & -- & 3.92 \\
\hline
\multicolumn{6}{l}{\textit{Standard NVAR}} \\
R1: East Sea       & 20 & --  & --                    & $1.56 \times 10^{5}$ & 0.05 \\
R2: Yellow Sea     & 20 & --  & --                    & $1.56 \times 10^{5}$ & 0.05 \\
R3: East China Sea & 20 & --  & --                    & $1.56 \times 10^{5}$ & 0.04 \\
\hline
\end{tabular}
\end{table}

Across all three oceanographic domains, Adaptive NVAR consistently selected the upper bound of the evaluated delay capacity ($k = 50$) and hidden layer dimensionality ($d_h = 100$), with learning rates settling between $4.15 \times 10^{-2}$ and $5.50 \times 10^{-2}$. Standard NVAR converged to a moderate delay length ($k = 20$) paired with substantial Tikhonov regularization ($\gamma = 1.56 \times 10^5$). 

The disparity in search execution times—approximately 3 minutes ($\sim 0.05$~h) for Standard NVAR versus 4 hours for Adaptive NVAR—reflects the iterative backpropagation overhead and the 5-run ensemble evaluation required per trial in the adaptive framework. However, this optimization time represents a one-time offline setup cost. Once trained, forward-pass inference across a 90-day forecast horizon executes in milliseconds, making the framework well-suited for real-time operational ocean forecasting.

\subsection{Experimental Results}
\label{Subsec:Experimental Results}

The multi-step forecasting capabilities of the Standard NVAR and Adaptive NVAR models were evaluated against a Persistence baseline across lead times of 10, 30, 60, and 90 days. Multi-step forecasts were generated recursively, with each predicted state fed back as the input for the next forecast step. The Persistence baseline assumed no temporal evolution by holding the initial observed SST field of each test window constant throughout the forecast horizon. This widely used benchmark provides a simple reference for assessing whether a model captures meaningful temporal dynamics. The comprehensive performance metrics, including MSE, MAE, and RMSE, alongside their relative improvements over the Persistence model, are detailed in Table~\ref{tab:sst_forecast_skill_landscape}.

For short-term prediction horizons (Step 10), the oceanic state exhibits strong temporal persistence due to its inherent thermal inertia, making the Persistence model a particularly challenging baseline. In the East Sea, Persistence marginally outperforms both approaches, with Adaptive NVAR producing an MSE of 0.0710, corresponding to an 8.54\% higher error than Persistence. This initial deficit represents an expected trade-off of the proposed framework. The East Sea is characterized by highly mesoscale dynamics and pronounced eddy activity, and reducing the high-dimensional SST field to a single principal component ($n_{\mathrm{comp}} = 1$) inevitably suppresses fine-scale, high-frequency spatial variability that is naturally preserved by the Persistence baseline. In addition, the hyperparameter optimization procedure was explicitly designed to minimize forecasting error at the 90-day horizon, biasing the models toward long-term stability rather than maximizing short-term accuracy. Consequently, a slight reduction in short-range predictive skill is expected. In contrast, for the comparatively less chaotic Yellow Sea and East China Sea, Adaptive NVAR immediately demonstrates superior predictive performance, achieving MSE reductions of 15.72\% and 17.15\%, respectively, relative to Persistence at Step 10. Standard NVAR likewise outperforms Persistence in both regions, although its improvements are consistently smaller than those obtained by the proposed Adaptive NVAR framework.

As the forecast horizon extends to medium- and long-term ranges (Steps 30, 60, and 90), the structural advantages of the Adaptive NVAR framework become highly pronounced. By Step 90, the Persistence baseline's error grows substantially across all domains. While Standard NVAR provides moderate long-term corrections—achieving maximum MSE improvements of 57.77\% in the East Sea, 65.66\% in the Yellow Sea, and 66.21\% in the East China Sea—it remains susceptible to continuous error accumulation typical of extended auto-regressive rollouts. 

In contrast, Adaptive NVAR successfully suppresses long-term error propagation, delivering exceptional Step 90 MSE improvements of 94.42\%, 96.52\%, and 91.29\% in the East Sea, Yellow Sea, and East China Sea, respectively.

\begin{figure}[t]
\centering
\includegraphics[width=\linewidth]{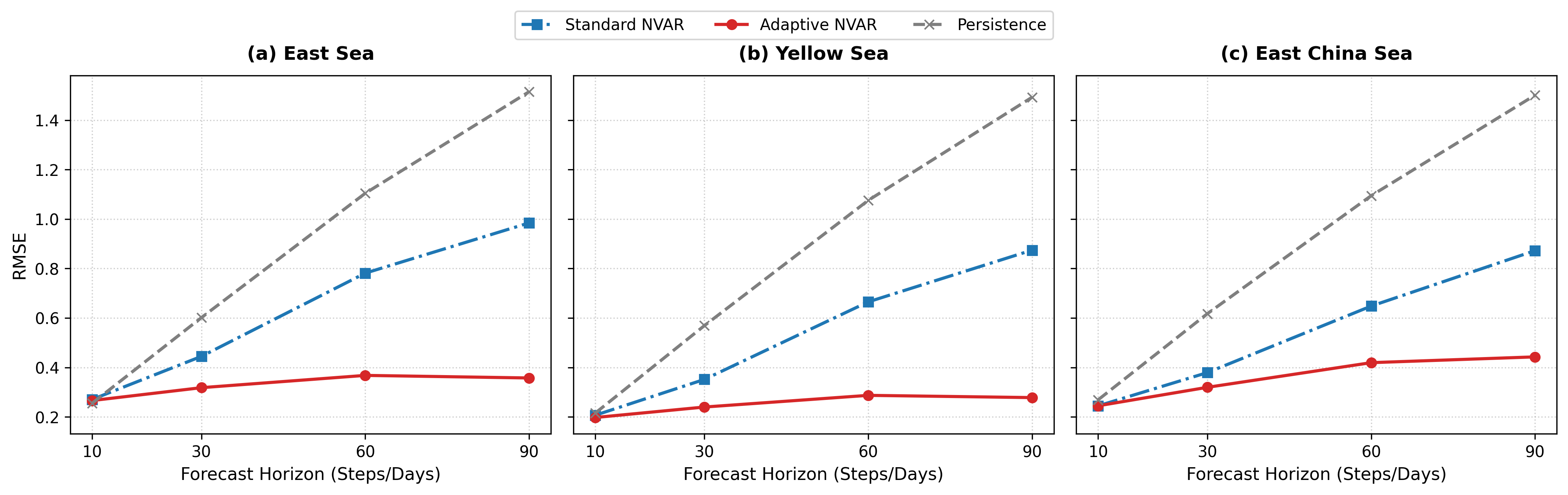}
\caption{Root Mean Square Error (RMSE) comparison of forecasting models across regional seas.}
\label{fig:sst_horizon_forecast_skill}
\end{figure}

This divergence in long-term stability is clearly illustrated in the horizon-wise error trajectories. As depicted in Figure~\ref{fig:sst_horizon_forecast_skill}, the RMSE for the Persistence baseline increases almost linearly with lead time. Standard NVAR exhibits a steadily climbing error curve, indicating gradual degradation in forecast reliability as time steps progress. Conversely, the Adaptive NVAR model establishes a remarkably flat error trajectory after the initial forecasting steps. It maintains consistent predictive skill and bounds the RMSE well below 0.45 across all three oceanographic regimes, even at the maximum 90-day horizon.

\begin{figure*}[htbp]
  \centering
  \begin{subfigure}[b]{0.49\textwidth}
    \centering
    \includegraphics[width=\textwidth]{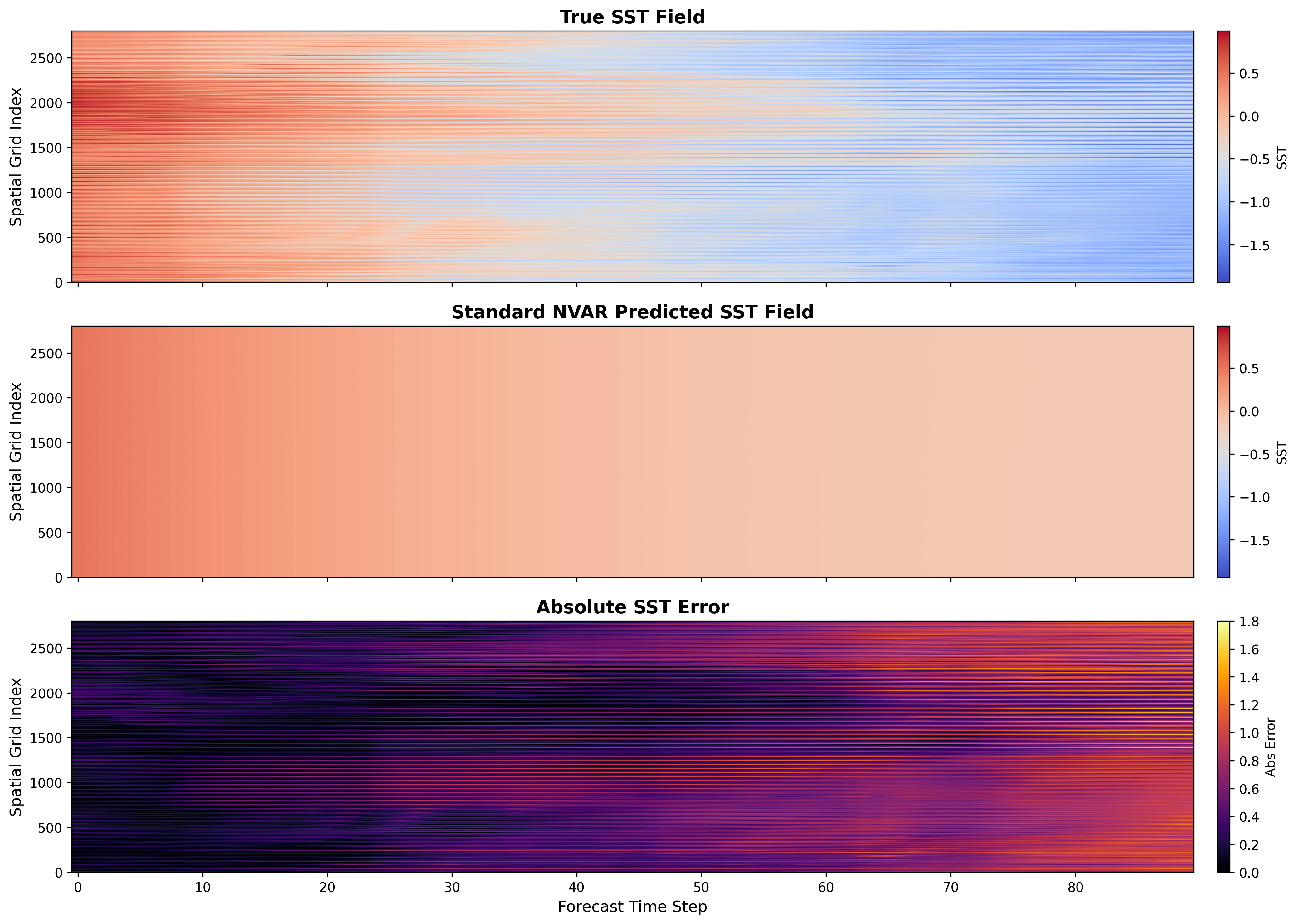}
    \caption{Standard NVAR}
    \label{fig:sst_east_sea_standard}
  \end{subfigure}
  \hfill
  \begin{subfigure}[b]{0.49\textwidth}
    \centering
    \includegraphics[width=\textwidth]{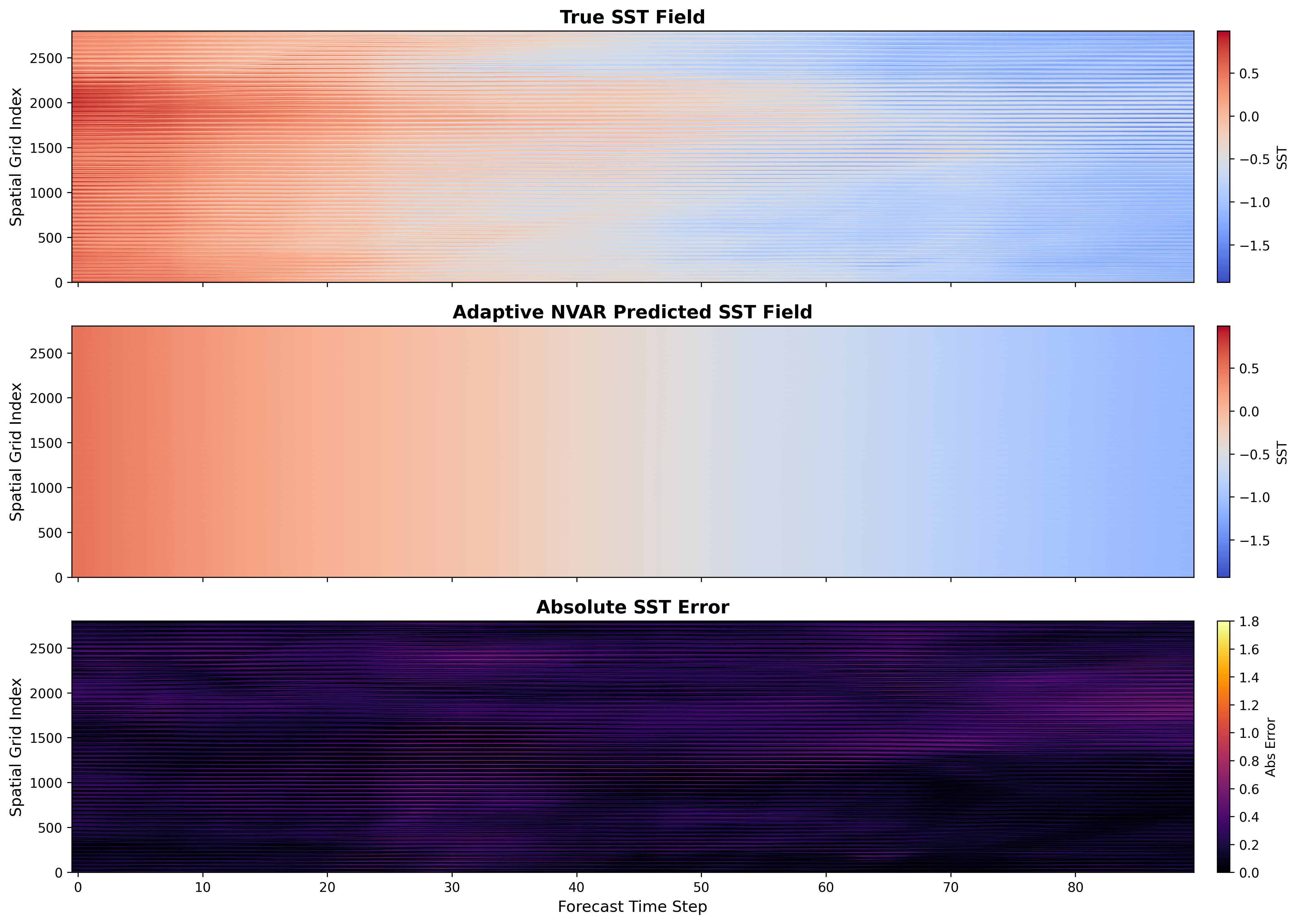}
    \caption{Adaptive NVAR Ensemble}
    \label{fig:sst_east_sea_adaptive}
  \end{subfigure}
  \caption{SST field evolution and absolute error comparison for the East Sea region: (a) Standard NVAR model and (b) Adaptive NVAR Ensemble model.}
  \label{fig:sst_east_sea_comparison}
\end{figure*}

\begin{figure*}[htbp]
  \centering
  \begin{subfigure}[b]{0.49\textwidth}
    \centering
    \includegraphics[width=\textwidth]{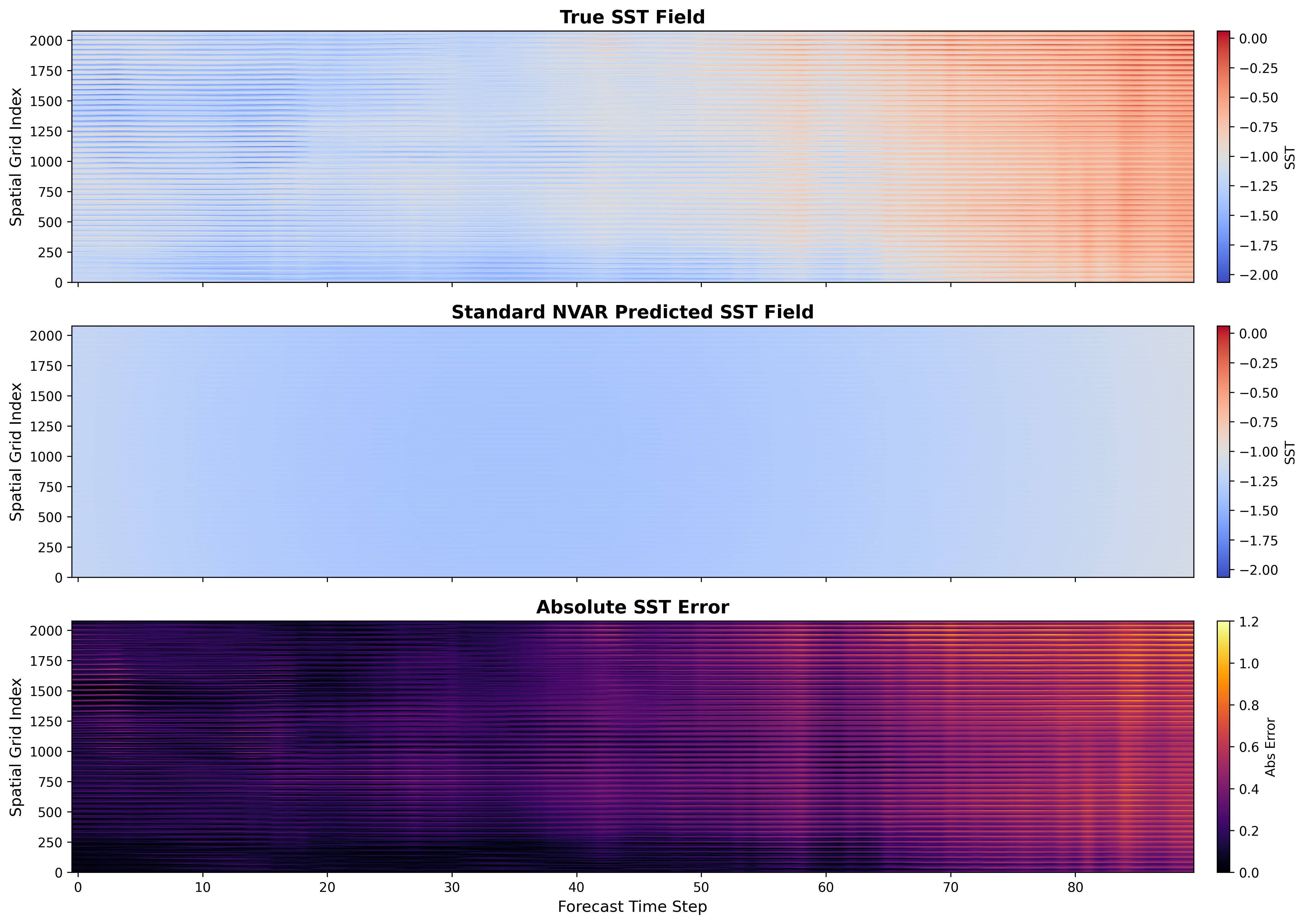}
    \caption{Standard NVAR}
    \label{fig:sst_yellow_sea_standard}
  \end{subfigure}
  \hfill
  \begin{subfigure}[b]{0.49\textwidth}
    \centering
    \includegraphics[width=\textwidth]{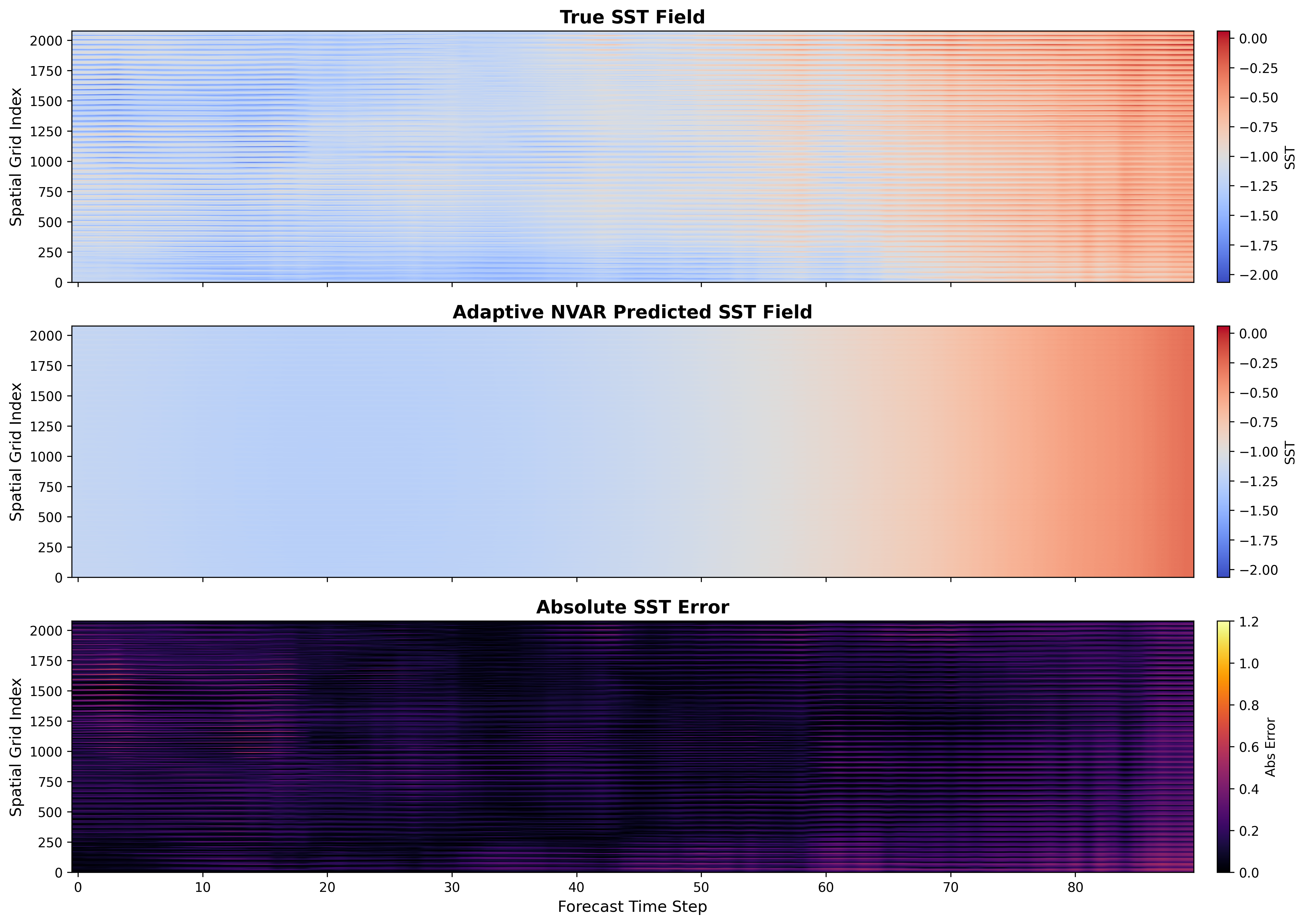}
    \caption{Adaptive NVAR Ensemble}
    \label{fig:sst_yellow_sea_adaptive}
  \end{subfigure}
  \caption{SST field evolution and absolute error comparison for the Yellow Sea region: (a) Standard NVAR model and (b) Adaptive NVAR Ensemble model.}
  \label{fig:sst_yellow_sea_comparison}
\end{figure*}

\begin{figure*}[htbp]
  \centering
  \begin{subfigure}[b]{0.49\textwidth}
    \centering
    \includegraphics[width=\textwidth]{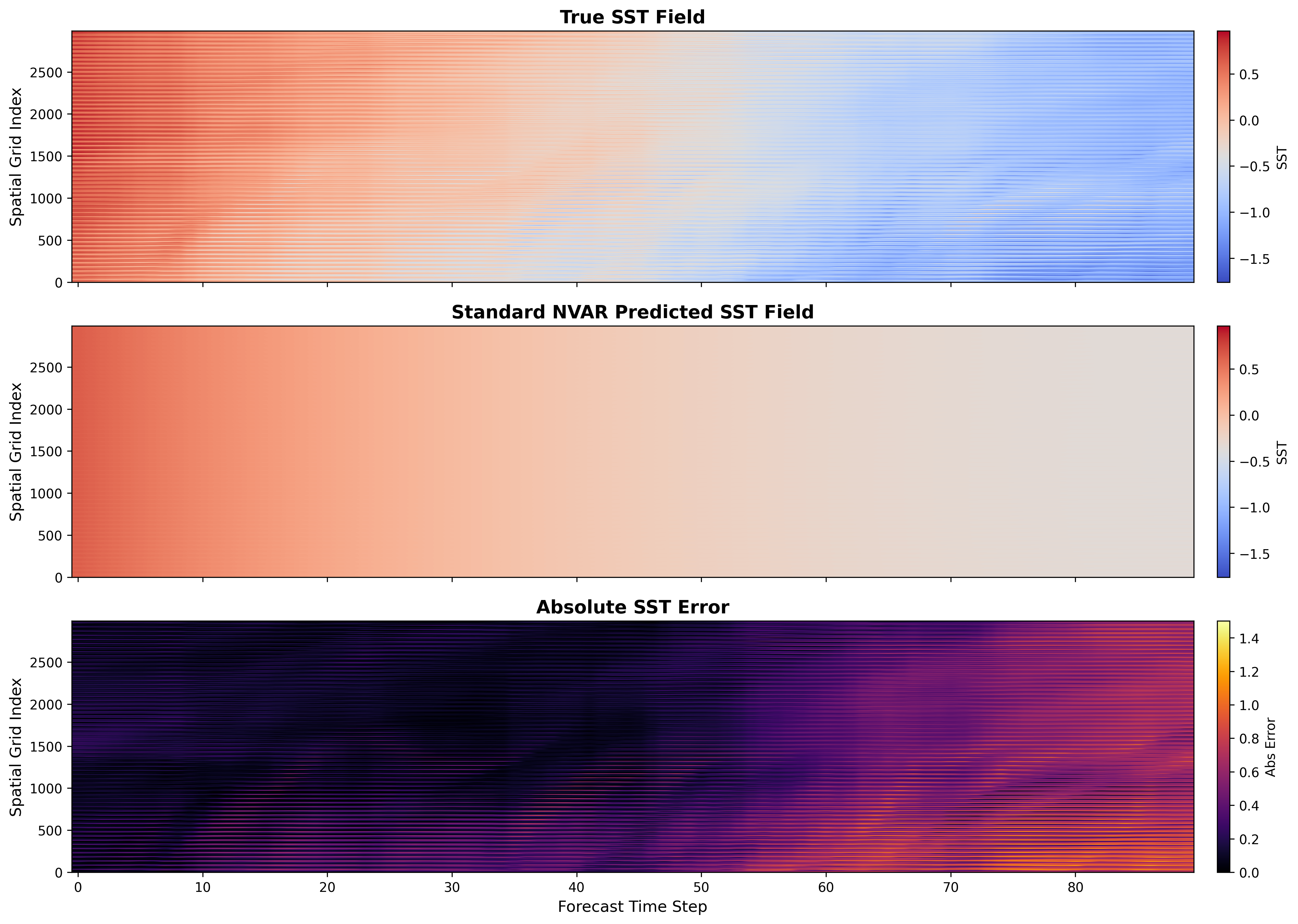}
    \caption{Standard NVAR}
    \label{fig:sst_east_china_sea_standard}
  \end{subfigure}
  \hfill
  \begin{subfigure}[b]{0.49\textwidth}
    \centering
    \includegraphics[width=\textwidth]{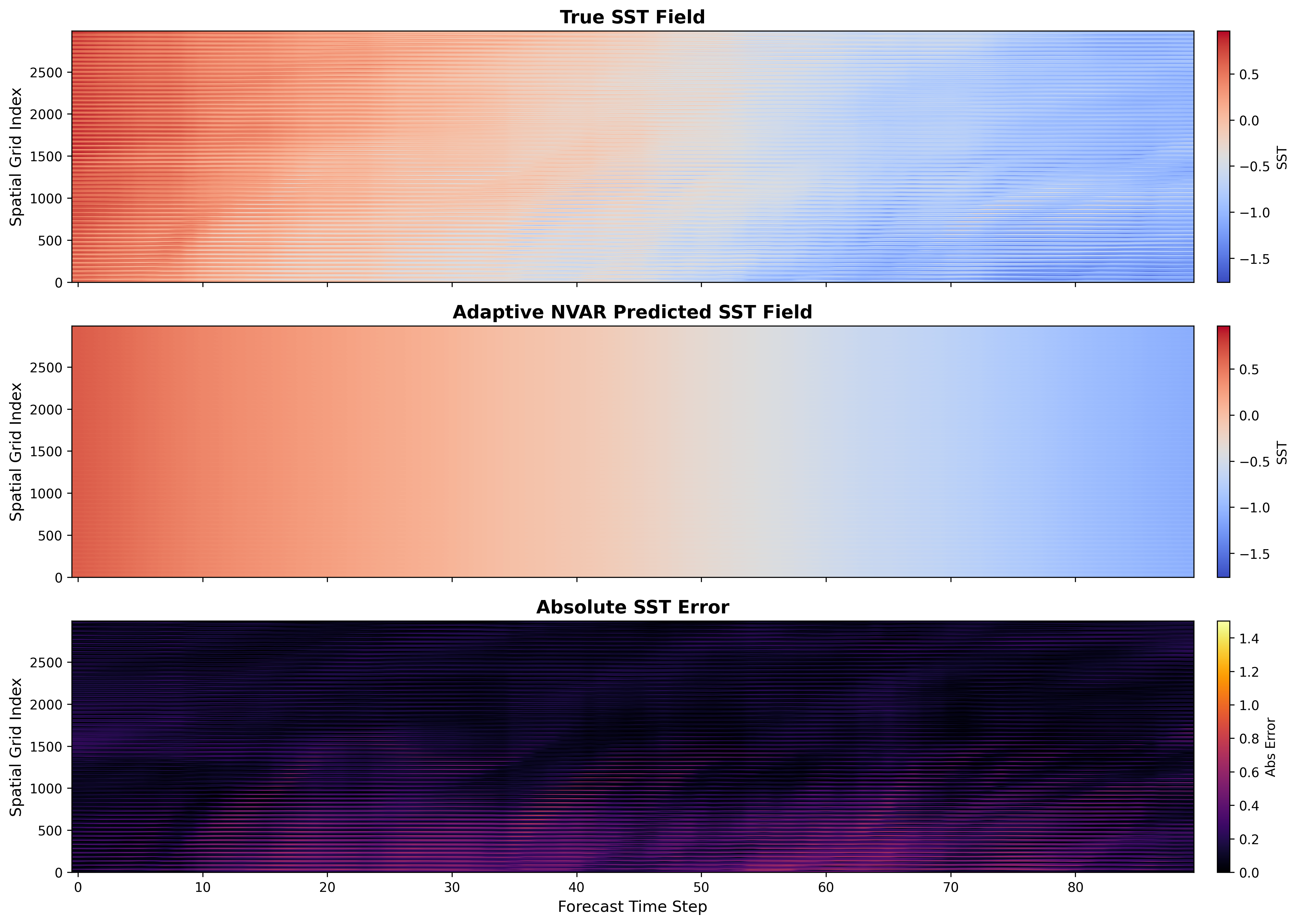}
    \caption{Adaptive NVAR Ensemble}
    \label{fig:sst_east_china_sea_adaptive}
  \end{subfigure}
  \caption{SST field evolution and absolute error comparison for the East China Sea region: (a) Standard NVAR model and (b) Adaptive NVAR Ensemble model.}
  \label{fig:sst_east_china_sea_comparison}
\end{figure*}

To further elucidate the spatio-temporal dynamics and the nature of the error accumulation, Figures~\ref{fig:sst_east_sea_comparison}, \ref{fig:sst_yellow_sea_comparison}, and \ref{fig:sst_east_china_sea_comparison} present a granular view of the SST field evolution and the corresponding absolute errors across the 90-day forecast horizon for the East Sea, Yellow Sea, and East China Sea, respectively. In these visualizations, the spatial grid indices are flattened along the y-axis, while the x-axis represents the forecast time steps, allowing for a comprehensive assessment of how spatial biases propagate over time.

The Standard NVAR (Panel (a) in all three figures) predictions suffer from severe error accumulation over time. As the forecast horizon extends toward 90 days, the predicted fields lose spatial details, and the absolute error maps show high-error bands, with values rising sharply up to $1.2$–$1.8$. This shows that the standard model drifts significantly from the true state.

In contrast, the Adaptive NVAR (Panel (b) in all three figures) Ensemble maintains high fidelity throughout the entire evaluation period. Its predicted SST fields closely match the true evolution, and the absolute error heatmaps remain consistently dark and uniform. This demonstrates that the adaptive framework successfully suppresses error growth and prevents long-term drift across all study regions.

\begin{sidewaystable}[htbp]
\centering
\caption{Horizon-wise forecast skill for reconstructed Sea Surface Temperature (SST) across three sea regions. Metrics compare Standard NVAR and Adaptive NVAR against Persistence across forecast horizons.}
\label{tab:sst_forecast_skill_landscape}
\scriptsize
\setlength{\tabcolsep}{3pt} 
\begin{tabular}{ll c rrr rrr rrr rrr}
\toprule
& & & \multicolumn{3}{c}{\textbf{Step 10}} & \multicolumn{3}{c}{\textbf{Step 30}} & \multicolumn{3}{c}{\textbf{Step 60}} & \multicolumn{3}{c}{\textbf{Step 90}} \\
\cmidrule(lr){4-6} \cmidrule(lr){7-9} \cmidrule(lr){10-12} \cmidrule(lr){13-15}
\textbf{Region} & \textbf{Model} & \textbf{Metric} & \textbf{Value} & \textbf{Persist.} & \textbf{Imp. (\%)} & \textbf{Value} & \textbf{Persist.} & \textbf{Imp. (\%)} & \textbf{Value} & \textbf{Persist.} & \textbf{Imp. (\%)} & \textbf{Value} & \textbf{Persist.} & \textbf{Imp. (\%)} \\
\midrule
\multirow{6}{*}{\textbf{East Sea}} 
 & \multirow{3}{*}{Standard NVAR} 
   & MSE  & 0.0735 & 0.0654 & -12.27 & 0.1981 & 0.3621 & +45.30 & 0.6107 & 1.2193 & +49.92 & 0.9691 & 2.2945 & +57.77 \\
 & & MAE  & 0.2132 & 0.2030 & -4.99  & 0.3592 & 0.5178 & +30.63 & 0.6483 & 0.9610 & +32.54 & 0.8356 & 1.2964 & +35.54 \\
 & & RMSE & 0.2711 & 0.2558 & -5.96  & 0.4451 & 0.6018 & +26.04 & 0.7815 & 1.1042 & +29.23 & 0.9844 & 1.5148 & +35.01 \\
\cmidrule(lr){2-15}
 & \multirow{3}{*}{Adaptive NVAR} 
   & MSE  & 0.0710 & 0.0654 & -8.54  & 0.1014 & 0.3621 & +71.99 & 0.1355 & 1.2193 & +88.89 & 0.1280 & 2.2945 & +94.42 \\
 & & MAE  & 0.2107 & 0.2030 & -3.79  & 0.2567 & 0.5178 & +50.42 & 0.2997 & 0.9610 & +68.82 & 0.2879 & 1.2964 & +77.79 \\
 & & RMSE & 0.2665 & 0.2558 & -4.18  & 0.3185 & 0.6018 & +47.07 & 0.3681 & 1.1042 & +66.66 & 0.3577 & 1.5148 & +76.38 \\
\midrule
\multirow{6}{*}{\textbf{Yellow Sea}} 
 & \multirow{3}{*}{Standard NVAR} 
   & MSE  & 0.0429 & 0.0465 & +7.78  & 0.1238 & 0.3241 & +61.80 & 0.4426 & 1.1569 & +61.74 & 0.7646 & 2.2269 & +65.66 \\
 & & MAE  & 0.1613 & 0.1719 & +6.19  & 0.2792 & 0.4797 & +41.79 & 0.5447 & 0.9195 & +40.76 & 0.7223 & 1.2716 & +43.20 \\
 & & RMSE & 0.2070 & 0.2156 & +3.97  & 0.3519 & 0.5693 & +38.19 & 0.6653 & 1.0756 & +38.14 & 0.8744 & 1.4923 & +41.40 \\
\cmidrule(lr){2-15}
 & \multirow{3}{*}{Adaptive NVAR} 
   & MSE  & 0.0392 & 0.0465 & +15.72 & 0.0577 & 0.3241 & +82.19 & 0.0826 & 1.1569 & +92.86 & 0.0775 & 2.2269 & +96.52 \\
 & & MAE  & 0.1540 & 0.1719 & +10.45 & 0.1924 & 0.4797 & +59.90 & 0.2302 & 0.9195 & +74.97 & 0.2300 & 1.2716 & +81.91 \\
 & & RMSE & 0.1979 & 0.2156 & +8.20  & 0.2403 & 0.5693 & +57.80 & 0.2875 & 1.0756 & +73.27 & 0.2784 & 1.4923 & +81.34 \\
\midrule
\multirow{6}{*}{\textbf{East China Sea}} 
 & \multirow{3}{*}{Standard NVAR} 
   & MSE  & 0.0601 & 0.0726 & +17.28 & 0.1439 & 0.3814 & +62.26 & 0.4203 & 1.1980 & +64.91 & 0.7611 & 2.2522 & +66.21 \\
 & & MAE  & 0.1825 & 0.2033 & +10.23 & 0.2965 & 0.5077 & +41.60 & 0.5353 & 0.9212 & +41.89 & 0.7260 & 1.2736 & +42.99 \\
 & & RMSE & 0.2451 & 0.2695 & +9.05  & 0.3794 & 0.6176 & +38.57 & 0.6483 & 1.0945 & +40.77 & 0.8724 & 1.5007 & +41.87 \\
\cmidrule(lr){2-15}
 & \multirow{3}{*}{Adaptive NVAR} 
   & MSE  & 0.0602 & 0.0726 & +17.15 & 0.1024 & 0.3814 & +73.16 & 0.1761 & 1.1980 & +85.30 & 0.1962 & 2.2522 & +91.29 \\
 & & MAE  & 0.1841 & 0.2033 & +9.45  & 0.2511 & 0.5077 & +50.55 & 0.3397 & 0.9212 & +63.13 & 0.3658 & 1.2736 & +71.28 \\
 & & RMSE & 0.2453 & 0.2695 & +8.98  & 0.3200 & 0.6176 & +48.19 & 0.4196 & 1.0945 & +61.66 & 0.4430 & 1.5007 & +70.48 \\
\bottomrule
\end{tabular}
\end{sidewaystable}

\section{Conclusion}
\label{Sec:Conclusion}
This study presented a reduced-order forecasting framework that integrates PCA with the Adaptive NVAR model for long-term autonomous forecasting of SST across the East Sea, Yellow Sea, and East China Sea. To enable efficient forecasting of high-resolution SST fields, the data were projected onto a one-dimensional latent space, where the leading principal component retained more than 96\% of the total variance while substantially reducing the computational complexity of the prediction task.
Experimental results demonstrated that Adaptive NVAR consistently outperformed the conventional Standard NVAR model and delivered superior overall forecasting performance relative to the Persistence baseline across the three study regions. Although the adaptive framework exhibited a modest reduction in short-term accuracy in the highly dynamic East Sea, it substantially reduced autoregressive error accumulation over extended prediction horizons. At the 90-day forecast horizon, Adaptive NVAR achieved MSE improvements exceeding 90\% relative to Persistence in all three regions while maintaining stable spatio-temporal SST evolution throughout the forecasting period. These results indicate that adaptive nonlinear feature learning provides a significantly more robust representation of long-term ocean dynamics than the fixed nonlinear mapping employed by standard NVAR (NG-RC).

The improved predictive performance is achieved at the expense of a higher one-time offline optimization cost. The optimal Adaptive NVAR configuration required approximately four hours of hyperparameter optimization, compared with approximately three minutes for Standard NVAR. However, once trained, both models perform autonomous 90-day forecasts within milliseconds, making the proposed framework well suited for real-time operational forecasting applications where offline training cost is incurred only once.

Despite these encouraging results, several limitations remain. First, the use of a single principal component ($n_{\mathrm{comp}}=1$) inevitably suppresses part of the fine-scale spatial variability associated with mesoscale ocean processes, which contributes to the slight reduction in short-term accuracy observed in the East Sea. Second, because the hyperparameter optimization objective emphasized the 90-day forecast horizon, the learned model was intentionally biased toward long-term stability rather than short-term precision. Finally, as a purely data-driven forecasting framework, Adaptive NVAR does not explicitly incorporate physical conservation laws, potentially limiting its physical consistency under substantially longer autonomous integrations.

Future work will investigate higher-dimensional latent representations to better preserve fine-scale spatial variability and incorporate multiple ocean depth layers to capture three-dimensional ocean dynamics. Extending the proposed framework to additional ocean basins and other geophysical forecasting problems also represents a promising direction for future research.

\section*{Computational Environment}\label{Sec:computational-environment}
All experiments and analyses were conducted using the Jupyter Notebook~\cite{JN}. All computations were performed on the system configuration summarized in Table~\ref{tab:environment}. The main libraries used in this study were Matplotlib~\cite{hunter2007matplotlib} for visualization, NumPy~\cite{np} and SciPy~\cite{Sci} for numerical computations, Optuna~\cite{akiba2019optuna} for hyperparameter search, and PyTorch~\cite{PyTorch} for implementing the models.

\begin{table}[h!]
\centering
\caption{Runtime environment used for the experiments.}
\label{tab:runtime_environment}
\renewcommand{\arraystretch}{1.25}

\begin{tabular}{lc}
\toprule
\textbf{Name} & \textbf{Configuration} \\
\midrule
CPU
& Intel(R) Xeon(R) Gold 6242 CPU @ 2.80GHz \\
GPU
& NVIDIA A100-PCIE-40GB \\
Memory
& 376.54 GB RAM \\
Operating System
& Linux 3.10.0-1062.el7.x86\_64 (64-bit) \\
\bottomrule
\end{tabular}
\label{tab:environment}
\end{table}

\section*{Code Availability}\label{Sec:code}
The source code developed and used for the PCA-Enhanced Adaptive Nonlinear Vector Autoregression (NVAR) framework presented in this study is publicly available at: \href{https://github.com/AzimovSherkhon/PCA-Enhanced-Adaptive-NVAR-Framework-for-High-Resolution-Sea-Surface-Temperature-Forecasting.git}{github.com/AzimovSherkhon/PCA-Enhanced-Adaptive-NVAR-Framework-for-High-Resolution-Sea-Surface-Temperature-Forecasting.git}

The repository contains the implementation of the PCA preprocessing, adaptive NVAR forecasting framework, comparison methods, data preprocessing procedures, evaluation routines, and example scripts. An example is provided to verify the basic operation of the implementation without requiring the complete oceanographic datasets. 

\section*{Acknowledgments}
This work was supported by a National Research Foundation of Korea (NRF) grant funded by the Korean Government (MSIT) (2022R1A5A1033624; RS-2023-00242528); the Global Learning \& Academic Research Institution for Master’s, Ph.D. students, and Postdocs (LAMP) Program of the National Research Foundation of Korea (NRF) grant, funded by the Ministry of Education (No. RS-2023-00301938); the Korea Institute of Marine Science \& Technology Promotion (KIMST) funded by the Ministry of Oceans and Fisheries, Korea (RS-2025-02217872); and the Glocal University 30 Project at Pusan National University through the Institute for Regional System \& Education in Busan Metropolitan City, funded by the Ministry of Education(MOE) and the Busan Metropolitan City, Republic of Korea.(2025-glocal-02-004-M432-09). 

Additionally, the work of S. López-Moreno was supported by the Korea National Research Foundation (NRF) grant funded by the Korean government (MSIT) (RS-2024-00406152), and the work of E. Dolores Cuenca was supported by the Korea National Research Foundation (NRF) grant funded by the Korean government (MSIT) (RS-2025-00517727).

\section*{Correspondence}
Correspondence and requests for materials should be addressed to Sangil Kim (email: sangil.kim@pusan.ac.kr) and Sherkhon Azimov (email: sherxonazimov94@pusan.ac.kr).

\bibliographystyle{IEEEtran}
\bibliography{ref}

@IEEEtranBSTCTL{IEEEexample:BSTcontrol,
  CTLuse_forced_etal       = "yes",
  CTLmax_names_forced_etal = "3",
  CTLnames_show_etal       = "3"
}

@article{jean2021copernicus,
  title={The Copernicus global 1/12 oceanic and sea ice GLORYS12 reanalysis},
  author={Jean-Michel, Lellouche and Eric, Greiner and Romain, Bourdall{\'e}-Badie and Gilles, Garric and Ang{\'e}lique, Melet and Marie, Dr{\'e}villon and Cl{\'e}ment, Bricaud and Mathieu, Hamon and Olivier, Le Galloudec and Charly, Regnier and others},
  journal={Frontiers in Earth Science},
  volume={9},
  pages={698876},
  year={2021},
  publisher={Frontiers Media SA}
}

@conference{JN,
	Author = {Thomas Kluyver and Benjamin Ragan-Kelley and Fernando P{\'e}rez and Brian Granger and Matthias Bussonnier and Jonathan Frederic and Kyle Kelley and Jessica Hamrick and Jason Grout and Sylvain Corlay and Paul Ivanov and Dami{\'a}n Avila and Safia Abdalla and Carol Willing},
	Booktitle = {Positioning and Power in Academic Publishing: Players, Agents and Agendas},
	Editor = {F. Loizides and B. Schmidt},
	Organization = {IOS Press},
	Pages = {87 - 90},
	Title = {Jupyter Notebooks -- a publishing format for reproducible computational workflows},
	Year = {2016}}

@inproceedings{akiba2019optuna,
  title={Optuna: A next-generation hyperparameter optimization framework},
  author={Akiba, Takuya and Sano, Shotaro and Yanase, Toshihiko and Ohta, Takeru and Koyama, Masanori},
  booktitle={Proceedings of the 25th ACM SIGKDD international conference on knowledge discovery \& data mining},
  pages={2623--2631},
  year={2019}
}

@article{bergstra2011algorithms,
  title={Algorithms for hyper-parameter optimization},
  author={Bergstra, James and Bardenet, R{\'e}mi and Bengio, Yoshua and K{\'e}gl, Bal{\'a}zs},
  journal={Advances in neural information processing systems},
  volume={24},
  year={2011}
}

@article{ju2022impacts,
  title={Impacts of seasonal and interannual variabilities of Sea Surface Temperature on its short-term deep-learning prediction model around the southern coast of Korea},
  author={Ju, Ho-Jeong and Chae, JEONG-YEOB and Lee, EUN-JOO and Kim, YOUNG-TAEG and Park, JAE-HUN},
  journal={The Sea Journal of the Korean Society of Oceanography},
  volume={27},
  number={2},
  pages={49--70},
  year={2022},
  publisher={The Korean Society Of Oceanography}
}

@article{zhang2021sea,
  title={Sea surface temperature prediction with memory graph convolutional networks},
  author={Zhang, Xiaoyu and Li, Yongqing and Frery, Alejandro C and Ren, Peng},
  journal={IEEE Geoscience and Remote Sensing Letters},
  volume={19},
  pages={1--5},
  year={2021},
  publisher={IEEE}
}

@article{gauthier2021next,
  title={Next generation reservoir computing},
  author={Gauthier, Daniel J and Bollt, Erik and Griffith, Aaron and Barbosa, Wendson AS},
  journal={Nature communications},
  volume={12},
  number={1},
  pages={5564},
  year={2021},
  publisher={Nature Publishing Group UK London}
}

@article{sherkhon2025adaptive,
  title={Adaptive Nonlinear Vector Autoregression: Robust Forecasting for Noisy Chaotic Time Series},
  author={Sherkhon, Azimov and Lopez-Moreno, Susana and Dolores-Cuenca, Eric and Lee, Sieun and Kim, Sangil},
  journal={arXiv preprint arXiv:2507.08738},
  year={2025}
}

@article{shahi2022prediction,
  title={Prediction of chaotic time series using recurrent neural networks and reservoir computing techniques: A comparative study},
  author={Shahi, Shahrokh and Fenton, Flavio H and Cherry, Elizabeth M},
  journal={Machine learning with applications},
  volume={8},
  pages={100300},
  year={2022},
  publisher={Elsevier}
}

@article{ren2026dctfm,
  title={DCTFM: A dynamic causal-temporal forecasting model for marine data},
  author={Ren, Weijie and Gao, Mingjian and Ma, Hongwei and Na, Xiaodong},
  journal={Ocean Engineering},
  volume={358},
  pages={125733},
  year={2026},
  publisher={Elsevier}
}

@book{lorenz1956empirical,
  title={Empirical orthogonal functions and statistical weather prediction},
  author={Lorenz, Edward N and others},
  volume={1},
  year={1956},
  publisher={Massachusetts Institute of Technology, Department of Meteorology Cambridge}
}

@article{hannachi2007empirical,
  title={Empirical orthogonal functions and related techniques in atmospheric science: A review},
  author={Hannachi, Abdel and Jolliffe, Ian T and Stephenson, David B and others},
  journal={International journal of climatology},
  volume={27},
  number={9},
  pages={1119--1152},
  year={2007},
  publisher={Chichester; New York: Wiley, 1989-}
}

@article{eckart1936approximation,
  title={The approximation of one matrix by another of lower rank},
  author={Eckart, Carl and Young, Gale},
  journal={Psychometrika},
  volume={1},
  number={3},
  pages={211--218},
  year={1936},
  publisher={Springer-Verlag}
}

@article{halko2011finding,
  title={Finding structure with randomness: Probabilistic algorithms for constructing approximate matrix decompositions},
  author={Halko, Nathan and Martinsson, Per-Gunnar and Tropp, Joel A},
  journal={SIAM review},
  volume={53},
  number={2},
  pages={217--288},
  year={2011},
  publisher={SIAM}
}

@Article{       np,
 title         = {Array programming with {NumPy}},
 author        = {Charles R. Harris and K. Jarrod Millman and St{\'{e}}fan J.
                 van der Walt and Ralf Gommers and Pauli Virtanen and David
                 Cournapeau and Eric Wieser and Julian Taylor and Sebastian
                 Berg and Nathaniel J. Smith and Robert Kern and Matti Picus
                 and Stephan Hoyer and Marten H. van Kerkwijk and Matthew
                 Brett and Allan Haldane and Jaime Fern{\'{a}}ndez del
                 R{\'{i}}o and Mark Wiebe and Pearu Peterson and Pierre
                 G{\'{e}}rard-Marchant and Kevin Sheppard and Tyler Reddy and
                 Warren Weckesser and Hameer Abbasi and Christoph Gohlke and
                 Travis E. Oliphant},
 year          = {2020},
 month         = sep,
 journal       = {Nature},
 volume        = {585},
 number        = {7825},
 pages         = {357--362},
 doi           = {10.1038/s41586-020-2649-2},
 publisher     = {Springer Science and Business Media {LLC}},
 url           = {https://doi.org/10.1038/s41586-020-2649-2}
}

@ARTICLE{Sci,
  author  = {Virtanen, Pauli and Gommers, Ralf and Oliphant, Travis E. and
            Haberland, Matt and Reddy, Tyler and Cournapeau, David and
            Burovski, Evgeni and Peterson, Pearu and Weckesser, Warren and
            Bright, Jonathan and {van der Walt}, St{\'e}fan J. and
            Brett, Matthew and Wilson, Joshua and Millman, K. Jarrod and
            Mayorov, Nikolay and Nelson, Andrew R. J. and Jones, Eric and
            Kern, Robert and Larson, Eric and Carey, C J and
            Polat, {\.I}lhan and Feng, Yu and Moore, Eric W. and
            {VanderPlas}, Jake and Laxalde, Denis and Perktold, Josef and
            Cimrman, Robert and Henriksen, Ian and Quintero, E. A. and
            Harris, Charles R. and Archibald, Anne M. and
            Ribeiro, Ant{\^o}nio H. and Pedregosa, Fabian and
            {van Mulbregt}, Paul and {SciPy 1.0 Contributors}},
  title   = {{{SciPy} 1.0: Fundamental Algorithms for Scientific
            Computing in Python}},
  journal = {Nature Methods},
  year    = {2020},
  volume  = {17},
  pages   = {261--272},
  adsurl  = {https://rdcu.be/b08Wh},
  doi     = {10.1038/s41592-019-0686-2},
}

@article{PyTorch,
  title={Automatic differentiation in PyTorch},
  author={Paszke, Adam and Gross, Sam and Chintala, Soumith and Chanan, Gregory and Yang, Edward and DeVito, Zachary and Lin, Zeming and Desmaison, Alban and Antiga, Luca and Lerer, Adam},
  year={2017}
}

@article{kim2010seasonal,
  title={Seasonal variation of upper layer circulation in the northern part of the East/Japan Sea},
  author={Kim, Taekyun and Yoon, Jong-Hwan},
  journal={Continental shelf research},
  volume={30},
  number={12},
  pages={1283--1301},
  year={2010},
  publisher={Elsevier}
}

@article{lee2006intermediate,
  title={Intermediate Water Formation: at the Japan/East sea Subpolar Front},
  author={Lee, Craig M and Thomas, Leif N and Yoshikawa, Yutaka},
  journal={Oceanography},
  volume={19},
  number={3},
  pages={110--121},
  year={2006},
  publisher={JSTOR}
}

@article{trusenkova2009dynamically,
  title={Dynamically Induced Anomalies of the Japan/East Sea Surface Temperature},
  author={Trusenkova, Olga and Lobanov, Vyacheslav and Kaplunenko, Dmitry},
  journal={Ocean and Polar Research},
  volume={31},
  number={1},
  pages={11--29},
  year={2009},
  publisher={Korea Institute of Ocean Science \& Technology}
}

@book{kantz2003nonlinear,
  title={Nonlinear time series analysis},
  author={Kantz, Holger and Schreiber, Thomas},
  year={2003},
  publisher={Cambridge university press}
}

@inproceedings{kingma2014adam,
  author       = {Diederik P. Kingma and
                  Jimmy Ba},
  editor       = {Yoshua Bengio and
                  Yann LeCun},
  title        = {Adam: {A} Method for Stochastic Optimization},
  booktitle    = {3rd International Conference on Learning Representations, {ICLR} 2015,
                  San Diego, CA, USA, May 7-9, 2015, Conference Track Proceedings},
  year         = {2015},
  url          = {http://arxiv.org/abs/1412.6980},
  bibsource    = {dblp computer science bibliography, https://dblp.org}
}

@article{hornik1989multilayer,
  title={Multilayer feedforward networks are universal approximators},
  author={Hornik, Kurt and Stinchcombe, Maxwell and White, Halbert},
  journal={Neural networks},
  volume={2},
  number={5},
  pages={359--366},
  year={1989},
  publisher={Elsevier}
}

@article{tikhonov1977solutions,
  title={Solutions of ill posed problems},
  author={Tikhonov, Andrey Nikolayevich},
  year={1977},
  publisher={John Wiley \& Sons}
}

@article{hunter2007matplotlib,
  title={The matplotlib user’s guide},
  author={Hunter, John and Dale, Darren},
  journal={Matplotlib 0.90. 0 user’s guide},
  volume={487},
  year={2007}
}

\end{document}